\documentclass{article}
\usepackage{ijcai26}

\usepackage{times}
\usepackage{soul}
\usepackage{url}
\usepackage[hidelinks]{hyperref}
\usepackage[utf8]{inputenc}
\usepackage[small]{caption}
\usepackage{graphicx}
\usepackage{amsmath}
\usepackage{amsthm}
\usepackage{booktabs}
\usepackage{algorithm}
\usepackage{algorithmic}
\usepackage[switch]{lineno}
\usepackage{xcolor}
\usepackage{amsfonts}

\usepackage{xspace}
\newcommand{\cham}{CHAM-net\xspace}
\newcommand{\teme}{TEM-E\xspace}
\newcommand{\temc}{TEM-C\xspace}
\newcommand{\fluxnete}{FLUXNET-E\xspace}
\newcommand{\fluxnetc}{FLUXNET-C\xspace}

\DeclareMathOperator*{\argmin}{arg\,min}

\usepackage{multirow}
\linenumbers

\newcommand{\squishlist}{
    \begin{list}{$\bullet$}
        { \setlength{\itemsep}{0pt}      \setlength{\parsep}{0pt}
            \setlength{\topsep}{0.5pt}       \setlength{\partopsep}{0pt}
            \setlength{\listparindent}{-2pt}
            \setlength{\itemindent}{-5pt}
            \setlength{\leftmargin}{0.5em} \setlength{\labelwidth}{0em}
            \setlength{\labelsep}{0.2em} } }
    
\newcommand{\squishend}{
\end{list}  }

\title{CHAM-net: A Contrastive Hierarchical Adaptive Meta-network for Robust Global Methane Flux Prediction}

\author{
Rongchao Dong$^1$
\and
Yiming Sun$^1$\and
Shuo Chen$^{2}$\and
Youmi Oh$^{3,4}$\and
Licheng Liu$^5$\and
Yiqun Xie$^6$\And
Xiaowei Jia$^1$\\
\affiliations
$^1$University of Pittsburgh\and 
$^2$Purdue University\and
$^3$University of Colorado Boulder\and
$^4$NOAA Global Monitoring Laboratory\and
$^5$University of Wisconsin–Madison\And
$^6$University of Maryland\\
\emails
\{rongchaodong, yis108, xiaowei\}@pitt.edu,
chen4371@purdue.edu, youmi.oh@noaa.gov, licheng.liu@wisc.edu, xie@umd.edu
}

\begin{document}

\maketitle

\begin{abstract}
Methane is a potent greenhouse gas that significantly contributes to global warming.
However, accurately estimating global methane emissions and consumption remains challenging due to the complex interactions among environmental drivers that may vary across spatial and temporal scales.
Prior data-driven methods often overlook the inherent spatiotemporal heterogeneity of ecosystems, failing to explicitly capture site-specific characteristics and cross-year evolutionary dynamics.
To address these issues, we propose the \textbf{C}ontrastive \textbf{H}ierarchical \textbf{A}daptive \textbf{M}eta-network (\textbf{CHAM-net}), a novel framework that explicitly learns from historical context to capture site-specific dynamics.
\cham employs a hierarchical encoder–decoder architecture, in which the encoder
captures site-specific characteristics from historical data and then dynamically conditions the decoder to generate the final prediction.
Experimental results demonstrate that \cham consistently outperforms all baseline methods on both simulation and observational datasets for methane emission and consumption, achieving nRMSE values as low as 0.43 and 0.88 with corresponding R$^2$ scores up to 0.97 and 0.68 for emission prediction.

\end{abstract}

\section{Introduction}
Methane (\texttt{$CH_4$}) is the second most significant greenhouse gas contributing to global warming after carbon dioxide ($CO_2$) and is responsible for about 30\% of the increase in global temperature since the industrial revolution
~\cite{second_gas}. Unlike carbon dioxide, methane is chemically reactive in the atmosphere and therefore has a relatively short atmospheric lifetime of about 9 years~\cite{lifetime}. This short lifetime means that reducing methane emissions can deliver rapid climate benefits, including slower warming rates, reduced climate extremes, and improved air quality \cite{benefit}, making methane mitigation one of the most effective near-term strategies for protecting human health, ecosystems, and vulnerable communities~\cite{ocko2021acting}. 

Traditional approaches primarily rely on process-based biogeochemistry models~\cite{teme,pb1,pb2} to simulate and estimate the natural methane cycle.
By incorporating the theoretical understanding of methane ecosystem dynamics with the key environmental drivers (e.g., soil features and temperatures), these models can be extended to enable global methane prediction and subsequent budget estimation.
However, they are often limited by rigid and extensive parameterization, leading to biased prediction and substantial computational requirements when applied over large regions and long time periods. 
Recently, data-driven machine learning (ML) methods~\cite{ml2,ml1,ml3,ml4} have emerged as a promising alternative, demonstrating strong capability in capturing non-linear relationships between environmental drivers and methane flux. 
Existing works have further explored knowledge transfer (e.g., pretraining and fine-tuning) from simulated data to sparse real observations (e.g., collected from eddy covariance towers) to improve the prediction accuracy ~\cite{XMethaneWet}.

\begin{figure}[!t]
    \centering
    \includegraphics[width=0.85\linewidth]{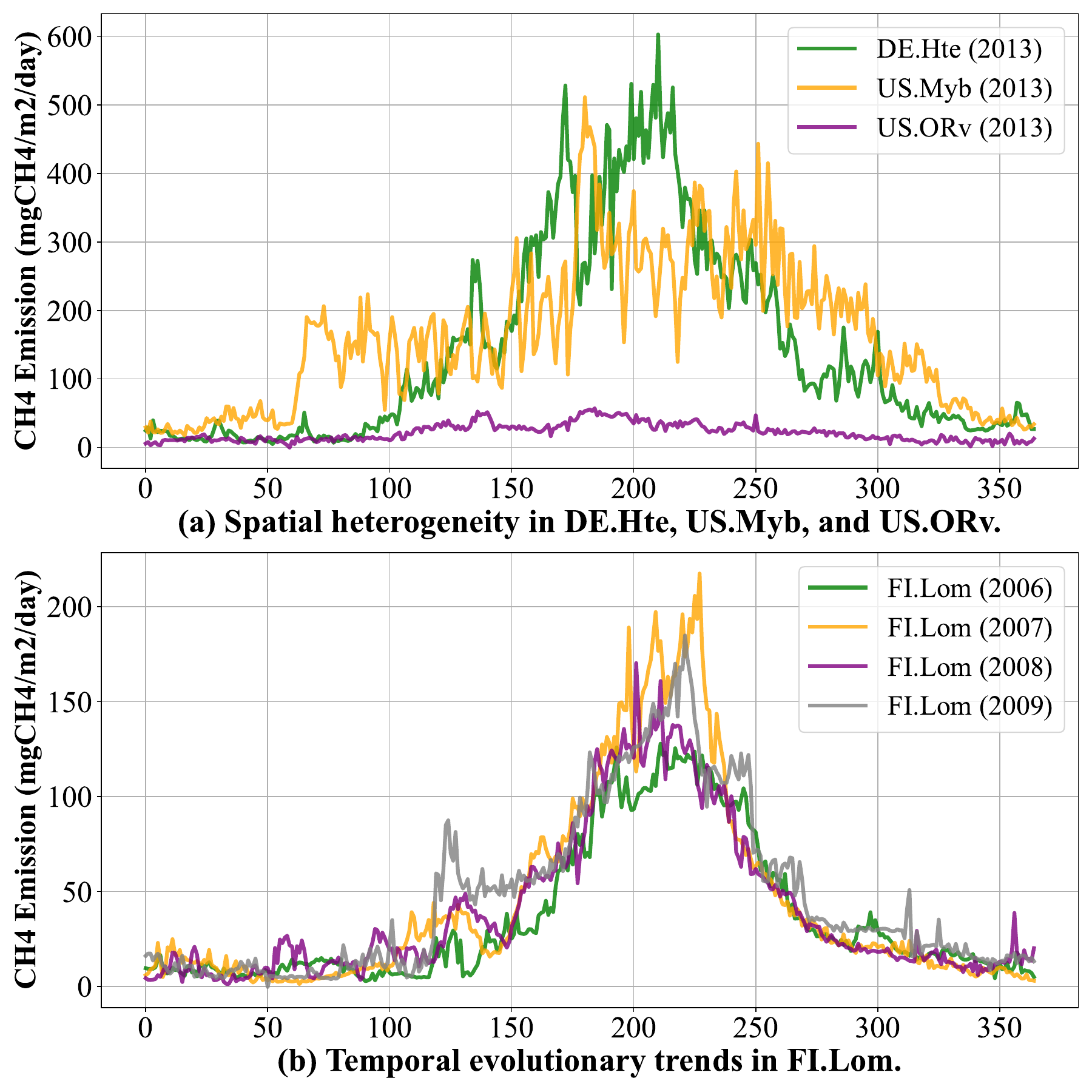}
    \vspace{-0.2cm}
    \caption{spatiotemporal heterogeneity within the datasets.}
    \vspace{-0.5cm}
    \label{fig:pattern_scale}
\end{figure}

However, existing ML methods typically assume 
diverse methane sites across space are 
governed by a shared set of global parameters, and neglect the spatiotemporal heterogeneity. 
In particular, methane emission and consumption patterns are highly site-specific because different sites can 
respond differently to similar input drivers due to the underlying variations in microbial communities and soil properties. For example, Figure \ref{fig:pattern_scale} (a) compares annual emissions at three sites (US.Myb~\cite{myb}, DE.Hte~\cite{hte}, and US.ORv~\cite{orv}) in 2013.  
These sites exhibit clearly different emission patterns and magnitudes. 
Without fully leveraging site-specific context, existing models tend to predict averaged values, which could underestimate high-valued regions and overestimate low-valued regions (e.g., the significant differences between sites DE.Hte and US.ORv in Figure \ref{fig:pattern_scale} (a)). 

Additionally, current ML methods mostly utilize short-term data (e.g., data from the current year) and focus on capturing short-term temporal dynamics (e.g., seasonal changes of precipitation). However, they are not designed to capture the impact of many long-term processes (e.g., slow changes in plant cover, soil composition) that also affect methane dynamics over years.  
As shown in Figure \ref{fig:pattern_scale} (b), 
for site FI.Lom~\cite{lom}, we can observe a sustained decline in total methane emissions from 2006 to 2009, reflecting a site-specific cross-year temporal evolution pattern.

To address these limitations, we propose the \textbf{C}ontrastive \textbf{H}ierarchical \textbf{A}daptive \textbf{M}eta-network (\textbf{\cham}), a novel framework that explicitly learns site-specific dynamics from multi-year historical data. 
While many site-specific characteristics (e.g., microbial communities and soil properties) are not directly observable, their influence is often manifested in the long-term trend shown in each site's historical methane record. \cham therefore leverages each site’s historical methane data to capture these latent characteristics and improve current-year prediction. 
Specifically, \cham adopts a hybrid meta-learning mechanism~\cite{hospedales2021meta} in which the inner model encodes multi-year environmental and methane dynamics into learnable representations to summarize site-specific characteristics (e.g., temporal patterns and scales), while the outer model leverages these learned representations to condition the current-year prediction process.
This design shifts prediction from a global model to a site-aware estimation. 
Additionally, optimizing the outer loop helps shape the inner-loop learning task, which enables effective extraction of site-specific information.  

\paragraph{Connections to the social good.} 
By applying our advanced AI architecture to resolve the spatiotemporal heterogeneity of methane ecosystems, this work significantly reduces uncertainties in natural methane budgets and improves process-level understanding, thereby enabling more effective and accurate methane mitigation strategies in support of the Global Methane Pledge~\cite{malley2023roadmap}. These advances directly contribute to multiple UN Sustainable Development Goals (https://sdgs.un.org/goals), including Climate Action, Good Health and Well-Being, and Life on Land, by informing near-term climate mitigation pathways and air-quality prediction, and supporting sustainable wetland management.

This paper is conducted in active collaboration with domain experts from 
NOAA Global Monitoring Laboratory, University of Wisconsin-Madison, University of Maryland, and Purdue University, 
who are internationally recognized experts in global methane observations, process understanding, and atmospheric data assimilation. These scientists contribute domain knowledge, observational constraints, and evaluate the real-world impact of these novel methodologies throughout model development, training, and interpretation.


\paragraph{Technical Contributions.} 

\squishlist{}
\item We identify inherent site heterogeneity as a key factor underlying the inaccurate predictions of the current models, and show that historical data encodes critical site-specific information essential for effective site prediction.
\item We propose \cham, an encoder–decoder hybrid meta-learning framework that dynamically leverages historical data to calibrate site-specific predictions for the current year.
\item We evaluate \cham on extensive methane emission and consumption datasets, including both simulation and observational data. Experimental results show that \cham consistently outperforms all baselines in all datasets, achieving an nRMSE of 0.88 and an R$^2$ of 0.68 in FLUXNET emission dataset.
\squishend{}

\section{Problem Formulation}
\label{back:problem}
The task of global methane flux prediction can be formulated as a site-level time-series regression problem. 
For each site $i\in \{1,\dots,N\}$, we are given a sequence of environmental drivers (e.g., soil properties and temperature) over a time period, denoted as  $X_i = \{x_1, x_2, \dots, x_T\}$, where each $x_t \in \mathbb{R}^{D}$ is a feature vector of dimension $D$ at time $t$ (e.g., a specific date), $D$ is the total number of input drivers, and $T$ is the length of each sequence. Following prior works in methane prediction and other environmental monitoring tasks~\cite{kgml3,XMethaneWet}, we cut the data into yearly sequences (i.e., $T$=365) to facilitate modeling seasonal patterns. The goal is to predict the methane flux for the corresponding year $Y_i = \{y_1, y_2, \dots, y_T\}$, where $y_t \in \mathbb{R}$ is the target label, e.g., methane emission or consumption.
In our proposed method, we also leverage multi-year historical records. For each site $i$, we use $X_i^{(k)}$ and $Y_i^{(k)}$ to represent the environmental drivers and methane data from a historical year $k\in \{1,\dots,K\}$.  


The datasets used in this paper can be categorized into simulation and observational datasets, each with distinct characteristics as follows:

\squishlist{}
\item \textbf{Simulation Dataset} We use process-based simulation datasets, 
which incorporate biogeochemical processes to simulate methane fluxes by solving differential equations. 
Given globally available input drivers, they enable estimation of methane fluxes at the global scale. However, due to the complexity of model computations and uncertainties in multiple input drivers, the highest spatial resolution is limited to 0.5 degree, corresponding to approximately 3,000 km$^2$ per grid.

\item \textbf{Observational Dataset} 
In observational datasets, both input drivers and methane fluxes are directly measured at each site using eddy covariance techniques, providing real-world ground truth observations.
However, these sites are spatially sparse and geographically discontinuous, with footprints on the order of hundreds of square meters.
Moreover, observational datasets cover substantially shorter time spans than simulation datasets.
The details of the datasets used in this paper are provided in Section \ref{datasets}.

\vspace{-0.2cm}
\section{Design and Methodology}
\label{design}
In this section, we introduce the main design of \cham (Contrastive Hierarchical Adaptive Meta-network) architecture, which addresses the aforementioned problems and leverages historical information to improve predictions.

\begin{figure*}[!h]
    \centering
    \includegraphics[width=0.7\linewidth]{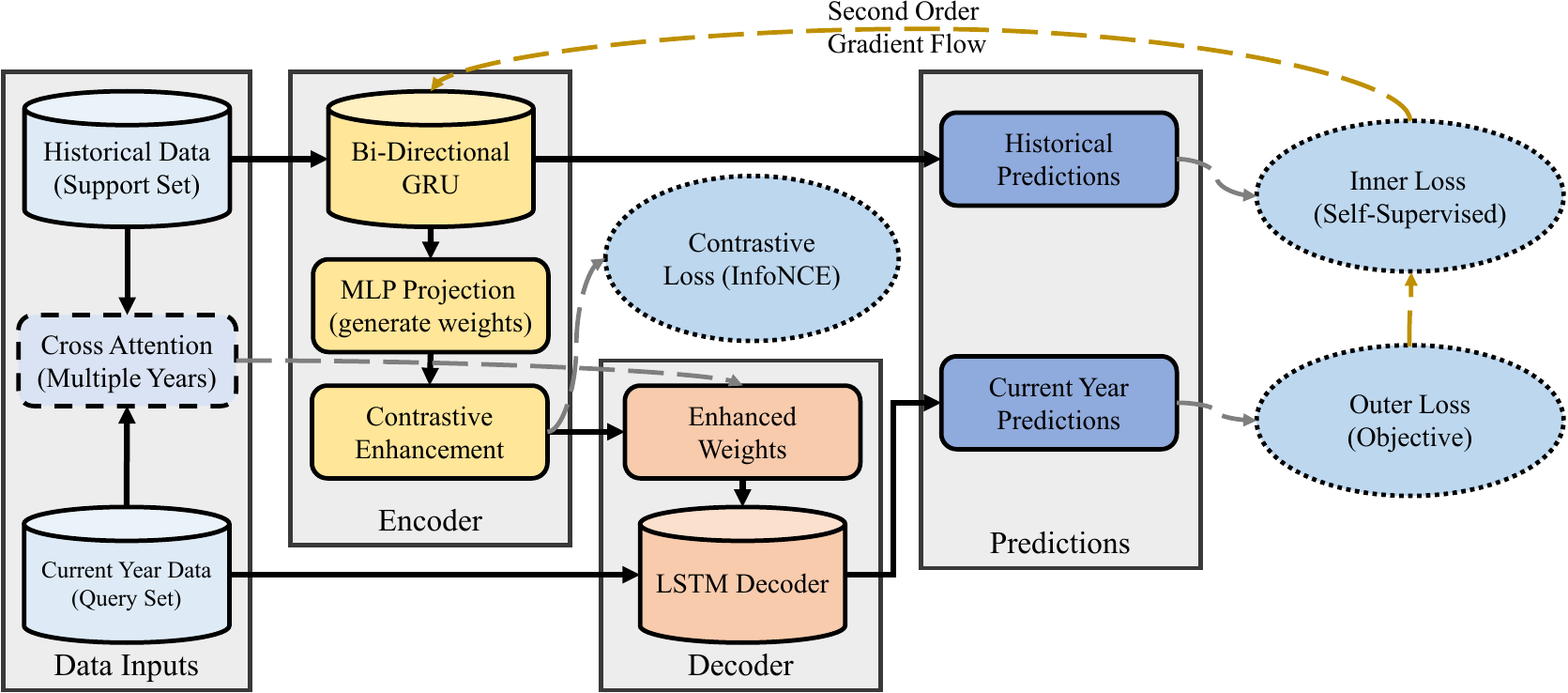}
    \vspace{-0.1cm}
    \caption{\cham structure overview.}
    \vspace{-0.5cm}
    \label{fig:model_structure}
\end{figure*}
\vspace{-0.1cm}
\subsection{Model Overview}
The proposed \cham model explicitly incorporates historical information to learn site-specific characteristics and long-term dynamics. 
By employing a hybrid hierarchical meta-learning architecture, the model extracts the most informative historical trends and scales for each site. These learned site-specific representations are then injected into the decoder and combined with current-year inputs to improve the final prediction. 

Figure \ref{fig:model_structure} illustrates the overall architecture of the \cham model. For each site, a configurable length of historical years' data is used as the \textbf{support set}, while the current year data forms the \textbf{query set}. 
When the historical year length exceeds one year, a cross-attention module is first applied to compute
each year's relevance to the current-year inputs. The attention weights are then used to extract context-aware representations capturing local characteristics that influence current-year dynamics. 
These representations are then propagated to the decoder stage to guide the final prediction.

The workflow consists of two hierarchical phases:
\squishlist{}
\item \textbf{Context Encoder.} 
The context encoder is designed to extract representations for capturing site-specific characteristics that affect current-year dynamics. Although many of these characteristics are not directly observable, they can be inferred from the dynamic responses of methane fluxes ${Y}$ to environmental drivers ${X}$, i.e., ${Y} = f({X};\theta)$.
The key idea of the context encoder is to \emph{inversely} infer these characteristics from historical methane data, by embedding them in a site-specific representation that serves as the parameters $\theta$ for the function $f$. 
However, directly inferring the $\theta$ as the full parameterization of the function $f$ may not precisely capture the site-specific characteristics as the mapping is also influenced by many confounding factors. Hence, we reformulate the model $f$ as ${Y} = f(g({X});\theta)$, where $g$ serves as a global feature extractor (via a bidirectional GRU) that captures shared underlying processes while $\theta$ parameterizes the site-specific information. 
The extractor $g$ is learned end-to-end under supervised training, which helps better define and stabilize the inverse problem for the target task.
A contrastive learning objective is further introduced to amplify inter-site differences, which ensures the discrimination of site-specific characteristics.



\item \textbf{Adaptive Decoder.} The resulting site-specific representations
are then transformed and injected into the hidden state of a Long Short-Term Memory (LSTM)~\cite{lstm}-based decoder.  
The decoder shares parameters across all sites,  but conditions its temporal dynamics on the injected site context. The final output of methane fluxes is produced from this context-enhanced hidden state.
\squishend{}

The model is trained through a bi-level optimization process. The inner-loop optimization aims to extract site-specific representations from historical data observations. In the outer loop,  the model utilizes these representations to condition and enhance the prediction, while updating model parameters to optimize predictive performance and to shape the inner-loop objective.  

In the following, we provide details of the model components as well as the training process. 

\vspace{-0.2cm}
\subsection{Context Encoder}
The goal of the encoder is to convert the historical support set $\mathcal{S}^{(k)} = \{(X_{i}^{t,(k)}, Y_{i}^{t,(k)})\}_{t=1}^{T}$ into a compact context-aware representation $\mathbf{W}_i$ for each site $i$, and then use it to enhance the decoder through site-specific conditioning. 

\textbf{BiGRU Module}. We first encode data from each historical year $k$ using a bidirectional GRU:

\begin{equation}
\mathbf{G}_i^{(k)} = \text{BiGRU}(\mathbf{X}_i^{(k)}) \in \mathbb{R}^{T \times 2H},
\end{equation}
where
$H$ is the size of the hidden dimension.

\textbf{Site-specific MLP}. To capture the site-specific characteristics, we assign each site $i$ a dedicated MLP projection head parameterized by $\mathbf{W}_i^k$, which can be represented by:
\begin{equation}
\hat{\mathbf{Y}}_i^{(k)} = \mathbf{G}_i^{(k)} \mathbf{W}_i^{k} + b_i,    
\end{equation}
where $\mathbf{W}_i^k \in \mathbb{R}^{2H \times 1}$ and $b_i \in \mathbb{R}$.
$\hat{\mathbf{Y}}_i^{(k)}$ represents the predicted labels of year $k$, which are later used to compute the inner-loop loss.

In our proposed method, we use the learned $\mathbf{W}_i^k$  as the representation of site-specific behavior.  
The MLP weights effectively distill rich historical information from both the drivers and methane fluxes into a compact representation. The dimensionality of $\mathbf{W}_i^k$ can be adjusted based on the output dimension of the previous GRU layer.


\textbf{Year-wise Attention}. When incorporating multiple historical years (i.e., $K > 1$), we weight each year based on its relevance to the current-year prediction. 
We therefore compute attention weights that measure the year-wise importance $\alpha_i^{(k)}$ of each historical year, as follows: 
\begin{equation}
\alpha_i^{(k)} =
\text{Attn}\big(
\mathbf{X}_i^{(0)},
\mathbf{X}_i^{(k)}
\big),
\quad
\text{s.t., \,}\sum_{k=1}^K \alpha_i^{(k)} = 1,    
\end{equation}
\vspace{-0.1cm}

where $\mathbf{X}_i^{(0)}$ is the current-year inputs. We employ the standard multiple-head attention mechanism to calculate the similarity between the current year and the historical years, and then normalize the attention weights across $K$ historical years via softmax. 
The attention weights are used in the outer-loop for aggregating $\mathbf{W}_i^k$ from historical years, which will be discussed in Section~\ref{decoder}. 

\textbf{Contrastive Learning}. 
To better ensure site discrimination and encode persistent site characteristics (e.g., long-term scaling and trend patterns) into $\mathbf{W}_i^k$, 
we introduce a contrastive objective directly defined over the site-specific representation $\mathbf{W}_i^k$. 
We treat the representation from the same site but different years as positive pairs $(\mathbf{W}_i^k, \mathbf{W}_i^+)$ and other sites in the batch as negatives. The contrastive loss is Info Noise-Contrastive Estimation (InfoNCE) loss, which is defined as:
\begin{equation}
\mathcal{L}_{cont} = - \log \frac{\exp(\text{sim}(\mathbf{W}_i^k, \mathbf{W}_i^+) / \tau)}{\sum_{j=1}^{B} \exp(\text{sim}(\mathbf{W}_i^k, \mathbf{W}_j) / \tau)},
\label{eq:cont_loss}
\end{equation}
where $B$ is the batch size, $\text{sim}(\cdot)$ is the cosine similarity, and $\tau$ is the configurable temperature parameter.

To improve the robustness of site-specific representations, we adopt a stochastic perturbation method to augment the generated $\mathbf{W}_i^k$ in contrastive learning:
\begin{equation}
\tilde{\mathbf{W}}_i^k = \mathbf{W}_i^k + \boldsymbol{\epsilon}, \quad
\boldsymbol{\epsilon}\sim \mathcal{N}(\mathbf{0}, \sigma^2\mathbf{I}),
\end{equation}

This augmentation encourages the embeddings to remain stable under small variations of site-specific parameters.

\vspace{-0.1cm}
\subsection{Adaptive Decoder}
\label{decoder}
The decoder serves as the base-learner that leverages the outcome of the encoder to condition the actual flux prediction on the Query Set $\mathcal{Q}_i = \{X_{curr}^{t}\}_{t=1}^{T}$.
In our implementation, the decoder is an LSTM-based model. 
We adopt LSTM because
LSTM-based architectures have consistently demonstrated robust performance in prior methane flux prediction studies~\cite{ml3,ml5,XMethaneWet}. 



We first 
aggregate $\tilde{\mathbf{W}}_i^k$ from multiple historical years using attention weights $\alpha_i^{(k)}$, 
and then project the site-specific embedding into the hidden state space of the decoder:
\begin{equation}
\tilde{\mathbf{z}}_i = \phi(\sum_{k=1}^K \alpha_i^{(k)} \, \tilde{\mathbf{W}}_i^k),
\end{equation}

where $\phi(\cdot)$ is a fully connected projection.

The $\tilde{\mathbf{z}}_i$ will be further normalized with the original hidden state of the LSTM decoder with a learnable weight $\beta_i$:
\begin{equation}
\tilde{\mathbf{h}}_i = 
\frac{\tilde{\mathbf{h}}_i + \beta_i \tilde{\mathbf{z}}_i}{ 1+|\beta_i|},
\end{equation}

Then the final prediction is obtained by putting the current-year inputs and the enhanced hidden state into the LSTM decoder:
\begin{equation}
\hat{\mathbf{Y}}_i^{(0)} =
\text{LSTM}_{\psi}
\big(
\mathbf{X}_i^{(0)}, \tilde{\mathbf{h}}_i
\big),
\end{equation}
where $\psi$ denotes the other parameters of the LSTM, and $\hat{\mathbf{Y}}_i^{(0)}$ is the predicted current-year labels.

\vspace{-0.1cm}
\subsection{Optimization and Loss Functions}
\label{sec:training}
The training of \cham is formulated as a bilevel optimization problem, involving nested inner- and outer-loop loss computations and the corresponding bilevel backpropagation process. The second-order backpropagation gradient flow is illustrated in Figure \ref{fig:model_structure}.

\textbf{Inner-loop Reconstruction Loss}. The inner-loop objective is designed to extract site representations using historical information. 
It consists of a historical reconstruction loss and a contrastive enhancement loss:
\begin{equation}
\mathcal{L}_{\text{inner}}(\theta)
=
\mathcal{L}_{\text{hist}}(\theta)
+
\lambda_{\text{con}} \mathcal{L}_{\text{con}}(\theta),
\end{equation}
where $\theta=\{\mathbf{W}_i^{k}\}_{k=1}^K$ is the parameters in the inner-loop, $\mathcal{L}_{\text{con}}(\theta)$ is the contrastive loss defined in Equation \ref{eq:cont_loss}, and $\lambda_{\text{con}}$ is a configurable hyperparameter that controls the contribution of the contrastive loss. 
The historical reconstruction loss $\mathcal{L}_{\text{hist}}(\theta)$ is defined as:
\begin{equation}
\mathcal{L}_{\text{hist}}(\theta)
=
\frac{1}{N}
\sum_{i=1}^N
\sum_{k=1}^K
\alpha_i^{(k)}
\left\|
\hat{\mathbf{Y}}_i^{(k)} - \mathbf{Y}_i^{(k)}
\right\|_2^2,
\end{equation}
where $N$ is the total number of sites in training.

The encoder parameters are updated through a differentiable inner optimization step:
\begin{equation}
\theta' = \theta - \gamma \nabla_\theta \mathcal{L}_{\text{inner}}(\theta),
\label{eq:inner_step}
\end{equation}
where $\gamma$ is the inner learning rate.

\textbf{Outer-loop Objective}. The outer-loop takes the learned representation  and performs current-year prediction, and the loss is calculated by the mean squared error across all the training sites:
\begin{equation}
\mathcal{L}_{\text{outer}}(\psi, \theta^*)
=
\frac{1}{N}
\sum_{i=1}^N
\left\|
\hat{\mathbf{Y}}_i^{(0)}(\theta', \psi)
-
\mathbf{Y}_i^{(0)}
\right\|_2^2
\end{equation}

The outer optimization updates the parameters $\psi$ for both the encoder and decoder parameters, i.e., the BiGRU parameters in the encoder and the LSTM parameters in the decoder. %
$\theta^*$ represents the updated site-specific representations obtained from the inner-loop (via Eq.~\ref{eq:inner_step}). 
Given that $\theta^*$ also contributes to the final prediction, the outer-loop errors are backpropagated to the representations $\theta$, 
which forms a second-order gradient flow during the training process.

The complete training objective is formulated as follows:
\begin{equation}
\begin{aligned}
\min_{\psi} \quad &
\mathcal{L}_{\text{outer}}(\psi, \theta^*) \\
\text{s.t.} \quad &
\theta^* = \argmin \mathcal{L}_\text{inner} (\theta)
\end{aligned}
\end{equation}
where $\psi$ is the parameters of the outer LSTM, $\theta$ is the parameters of inner-loop. Note that the outer-loop parameters $\psi$ are shared across all sites while the inner-loop parameters $\theta$ are specific to each site. 

\textit{\textbf{Why meta-learning}} 
Traditional methane flux prediction models either rely on globally shared parameters or fail to fully exploit historical information. 
In our setting, historical context must be explicitly encoded into the prediction process.
However, since input drivers and labels vary dynamically across training and evaluation, efficiently integrating this information poses a challenge. Meta-learning provides a natural solution. 
Each site can be viewed as a distinct and evolving task, characterized by unique long-term dynamics and temporal responses. 
Through the inner-loop, the model learns site-specific historical information and dynamically adapts it to the outer-loop objectives, offering an effective mechanism for improving the prediction accuracy.

\vspace{-0.2cm}
\section{Evaluation}

\subsection{Experimental Setup}
\label{datasets}
\textbf{Baselines.} We compare \cham against nine competitive long-term time-series forecasting models. 
These include Transformer-based methods, such as the original Transformer~\cite{transformer}, iTransformer~\cite{itransformer}, Pyraformer~\cite{pyraformer}, DUET~\cite{duet}, and PatchTST~\cite{patchtst}, as well as two MLP-based architectures, TSMixer~\cite{tsmixer} and TimeMixer~\cite{timemixer}.
We further include two RNN-based approaches, namely LSTM~\cite{lstm} and P-sLSTM~\cite{pslstm}. Notably, \cham can also be regarded as an RNN-based model, as it relies on BiGRU and LSTM components to extract historical information and condition the prediction.

\textbf{Datasets.} 
We evaluate \cham and all baseline models using both simulation and observational datasets. We also consider both methane emission and consumption processes, which together constitute the natural methane cycle.
Details of the datasets used in our experiments are provided below.

\squishlist{}
\item \textit{Emission Datasets}. We leverage the methane emission datasets introduced in~\cite{XMethaneWet}, which include a global simulation dataset at 0.5 degree spatial resolution with daily granularity, TEM~\cite{teme} (\textbf{TEM-E}), and an observational dataset derived from eddy covariance measurements, FLUXNET-CH4~\cite{fluxnet} (\textbf{FLUXNET-E}).
The TEM-E dataset provides 40 years of global data spanning from 1979 to 2018. The FLUXNET-E dataset consists of measurements from 30 wetland eddy covariance tower sites, with site-specific temporal coverage determined by each site,
but primarily ranging from 2006 to 2018.
For each site in both TEM-E and FLUXNET-E, the datasets include 15 methane-related scalar input drivers at daily resolution, along with daily methane emission fluxes as target labels. 
Among these inputs, 10 are static or multi-years evolving site-specific features, including elevation (\texttt{clelev}), soil texture fractions (sand, silt, and clay) (\texttt{clfaotxt}), vegetation type (\texttt{cltveg}), soil pH value (\texttt{phh2o}), topsoil bulk density (\texttt{topsoil\_bulk\_density}), plant functional type (\texttt{vegetation\_type\_11}), wetland type (\texttt{wetlandtype}), climate type (\texttt{climatetype}), atmospheric carbon dioxide concentration (\texttt{kco2}), and atmospheric methane concentration (\texttt{ch4}).
The remaining five inputs are dynamic climate-related variables, including precipitation (\texttt{PREC}), air temperature (\texttt{TAIR}), solar radiation (\texttt{SOLR}), vapor pressure (\texttt{VAPR}), and net primary productivity (\texttt{NPP}).
\item \textit{Consumption Datasets}.
We use two consumption simulation datasets, \textbf{\temc}~\cite{temc} and \textbf{MeMo}~\cite{memo2,memo1}, which are generated by different physics-based methane consumption models that incorporate different physical processes and geographic constraints.
We additionally include an observational dataset, \textbf{\fluxnetc}~\cite{fluxnet}, which consists of methane consumption measurements from 28 upland sites collected using eddy covariance techniques.
Both \temc and MeMo provide global coverage at 0.5 degree spatial resolution with monthly granularity data. The \temc dataset spans from 1979 to 2019, while MeMo covers the period from 1990 to 2009, yielding 20 years of data. 
The temporal coverage of \fluxnetc varies across sites but primarily ranges from 2006 to 2018.
Due to difficulties of real-world data collection, \fluxnetc contains a limited number of missing observations (e.g., NaN consumption values in certain months for some sites). In our experiments, these missing entries are excluded from loss computation, such that predictions corresponding to NaN labels do not contribute to the optimization objective.
Across all simulation and observational consumption datasets, we use a total of 18 input features. These include all drivers described in the emission datasets, along with two additional soil texture variables (\texttt{sand} and \texttt{clay}) and leaf area index (\texttt{LAI}), which characterizes vegetation cover in upland ecosystems.
\squishend{}

\textbf{Implementation Details}
All models are implemented in PyTorch 2.5.1 with CUDA 12.4. 
We use a learning rate of 0.001, a dropout rate of 0.2, a hidden dimension 8, three layers, 128 batch size for simulation datasets, 4 batch size for observational datasets, and the Gaussian Error Linear Unit (GELU) as the activation function~\cite{gelu} among all models. For all Transformer-based models, the number of attention heads is set to 4. Other model-specific hyperparameters follow the default settings provided in the original implementations or corresponding papers.
We also employ an early stopping strategy with a patience of five epochs. 
Each experiment is repeated three times, and the reported results are averaged across runs. All experiments are conducted using the Adam optimizer~\cite{adam} on a single NVIDIA GTX 3080 GPU.
We provide our datasets and code in the open-source Zenodo repository\footnote {Dataset and code link: https://zenodo.org/records/20450697.}.

\squishend{}

\textbf{Data Splitting and Evaluation Task.} 
We focus on temporal extrapolation for both methane emission and consumption tasks. Following~\cite{XMethaneWet}, simulation datasets are split chronologically into two equal halves, with the earlier period used for training and the later period used for testing. For example, in both the \teme and \temc datasets, data from 1979 to 1998 are used for training, while data from 1999 to 2018 are used for testing. 
For the observational datasets \fluxnete and \fluxnetc, data are split on a per-site basis, where six-sevenths of the temporal records are used for training and the remaining one-seventh for testing. This 6/7–1/7 split ensures sufficient training data for effective model learning while mitigating overfitting risks.

\textbf{Evaluation Metrics.} We use normalized root of mean squared error (nRMSE) and the coefficient of determination (R$^2$) to evaluate model performance. 
nRMSE measures the magnitude of prediction errors normalized by the scale of the observations, enabling fair comparison across sites with substantially different magnitudes. Lower nRMSE indicates better performance.
This metric particularly fits in our setting, where methane fluxes vary widely across locations. 
R$^2$ quantifies the proportion of variance in the observations explained by the model, reflecting its ability to capture the temporal patterns. Higher R$^2$ values correspond to a better model fit.
Together, nRMSE and R$^2$ provide complementary perspectives on predictive accuracy and pattern fidelity.



\vspace{-0.2cm}
\subsection{Results}
\subsubsection{Predictive performance}
Table \ref{main_results} reports the predictive performance on all simulation and observational datasets. For each column, models are trained and evaluated solely on the corresponding dataset, without pretraining or fine-tuning across datasets.
We observe that \cham consistently outperforms all baselines in terms of both nRMSE and R$^2$.
Specifically, on the emission datasets \teme and \fluxnete, \cham achieves nRMSE values of 0.43 and 0.88, respectively. The nRMSEs are at least 0.1 lower than those of competing methods, indicating its superior ability to capture the magnitude of methane emissions over time. \cham also attains the highest R$^2$ value on \fluxnete, showing more accurate modeling of the emission patterns.
The combination of lower nRMSE and higher R$^2$ demonstrates that \cham effectively captures both emission scales and temporal dynamics, which is critical for reliable future methane budget estimation.
Moreover, the improvement reflects that the learned site-specific representations provide valuable insights into  localized emission behavior, which can help inform 
targeted methane mitigation strategies.

For the consumption datasets, \cham demonstrates substantial improvements on \temc, reducing nRMSE by more than 0.4 and increasing R$^2$ by over 0.2, indicating a significant performance gain. For the MeMo dataset, which is generated using relatively simpler biogeochemical equations and has more regular patterns, nearly all models achieve lower nRMSE and higher R$^2$ compared to the \temc dataset.
Results on both \temc and MeMo demonstrate that explicitly leveraging historical information substantially enhances predictive performance for current-year consumption. 
For the observational dataset \fluxnetc, model performance is generally limited due to the small magnitude of consumption fluxes and high levels of environmental noise. 
Nevertheless, \cham still achieves the highest R$^2$ value of 0.31 and the lowest nRMSE of 1.27 among all methods.

\begin{table*}[!h]
    \centering
    \small
    \caption{Main results across all datasets and models. Each dataset is trained and evaluated from scratch. Lower nRMSE and higher R$^2$ indicate better performance.}
    \vspace{-.1in}
    \begin{tabular}{l|c|c|c|c|c|c|c|c|c|c|}
        \toprule
        \multirow{2}{*}{Methods} & \multicolumn{2}{c|}{TEM-E} & \multicolumn{2}{c|}{FLUXNET-E} & \multicolumn{2}{c|}{TEM-C} & \multicolumn{2}{c|}{MeMo} & \multicolumn{2}{c|}{FLUXNET-C} \\
        & nRMSE & R$^2$ & nRMSE & R$^2$ & nRMSE & R$^2$ & nRMSE & R$^2$ & nRMSE & R$^2$\\
        \midrule
        LSTM         & 0.57 & 0.94 & 1.08 & 0.53 & 1.56 & 0.56 & 0.22 & 0.89 & 1.31 & 0.27    \\
        P-sLSTM      & 1.34 & 0.71 & 1.06 & 0.54 & 1.82 & 0.41 & 0.30 & 0.82 & 1.41 & 0.16   \\
        Transformer  & 0.52 & 0.95 & 1.09 & 0.52 & 1.45 & 0.62 & 0.20 & 0.91 & 1.40 & 0.16    \\
        iTransformer & 1.24 & 0.75 & 1.29 & 0.33 & 1.64 & 0.52 & 0.29 & 0.83 & 1.36 & 0.21   \\
        Pyraformer   & 0.92 & 0.87 & 1.05 & 0.56 & 1.49 & 0.60 & 0.22 & 0.89 & 1.29 & 0.29   \\
        TSMixer      & 0.55 & 0.95 & 0.99 & 0.60 & 1.47 & 0.61 & 0.24 & 0.88 & 1.35 & 0.23   \\
        TimeMixer    & 1.85 & 0.45 & 1.56 & 0.02 & 1.79 & 0.42 & 0.30 & 0.82 & 1.48 & 0.07   \\
        PatchTST     & 1.81 & 0.48 & 1.26 & 0.36 & 1.93 & 0.33 & 0.43 & 0.62 & 1.38 & 0.20   \\
        DUET        & 1.62 & 0.58 & 1.18 & 0.44 & 1.81 & 0.41 & 0.29 & 0.83 & 1.38 & 0.19    \\
        \textbf{CHAM-net}     & \textbf{0.43} & \textbf{0.97} & \textbf{0.88} & \textbf{0.68} & \textbf{0.99} & \textbf{0.82} & \textbf{0.15} & \textbf{0.95} & \textbf{1.27} & \textbf{0.31}    \\
        \bottomrule
    \end{tabular}
    \vspace{-0.2cm}
    \label{main_results}
\end{table*}

\subsubsection{Case Analysis}
\begin{figure}[!h]
    \centering
    \includegraphics[width=0.85\linewidth]{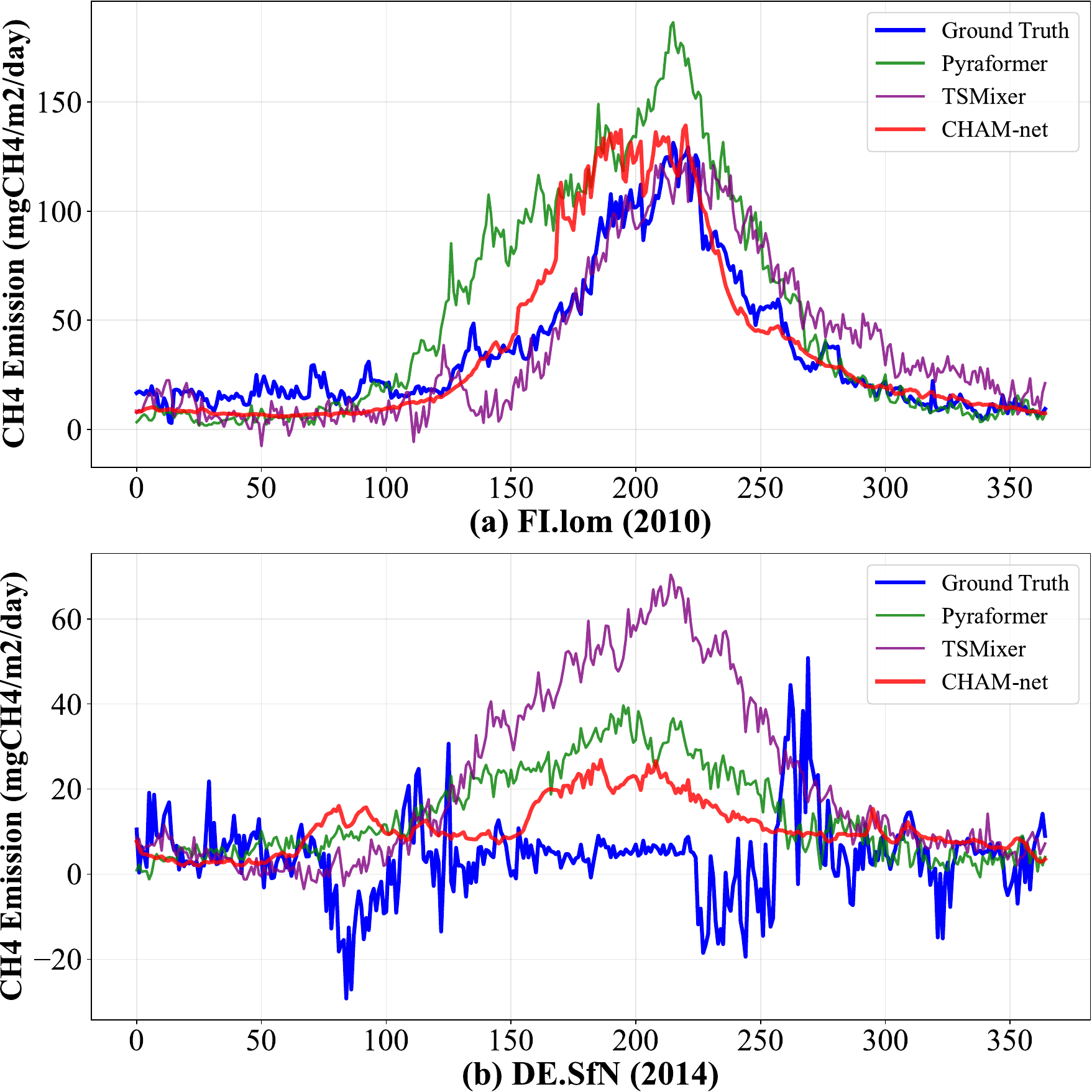}
    \vspace{-0.2cm}
    \caption{Predictions on representative \fluxnete sites.}
    \vspace{-0.5cm}
    \label{fig:case}
\end{figure}

Figure \ref{fig:case} presents two representative site examples from the \fluxnete dataset. We compare \cham with the two best baseline models, Pyraformer and TSMixer.
As shown in Figure \ref{fig:case} (a), \cham achieves a closer match to both the scale and temporal evolution of methane emissions at site FI-Lom, particularly after day 200. 
While all three models capture the overall temporal pattern, \cham more accurately estimates the peak value. In contrast, Pyraformer overestimates the peak value, whereas TSMixer underestimates it.
Figure \ref{fig:case} (b) illustrates results for site DE-SfN~\cite{desfn}, which represents a particularly challenging case due to the presence of negative flux values in the ground truth. 
Even under this difficult setting, \cham more effectively captures the site-specific emission scale than the baseline models, resulting in higher R$^2$, lower nRMSE, and a total emission estimate that is closer to the observed values.

\vspace{-0.3cm}

\begin{table*}[!htbp]
    \centering
    \caption{Performance of models pretrained on simulation datasets and fine-tuned on observational datasets.}
    \begin{tabular}{l|c|c|c|c|c|c|}
        \toprule
        \multirow{2}{*}{Methods} & \multicolumn{2}{c|}{\teme} & \multicolumn{2}{c|}{\temc} & \multicolumn{2}{c|}{MeMo} \\
         & nRMSE & R$^2$ & nRMSE & R$^2$ & nRMSE & R$^2$ \\
        \midrule
        LSTM         & 1.05 & 0.56 & 1.30 & 0.29 & 1.34 & 0.24    \\
        P-sLSTM      & 1.17 & 0.45 & 1.30 & 0.28 & 1.40 & 0.17    \\
        Transformer  & 1.09 & 0.52 & 1.30 & 0.29 & 1,32 & 0.27     \\
        iTransformer & 1.09 & 0.53 & 1.33 & 0.25 & 1.42 & 0.15   \\
        Pyraformer   & 1.07 & 0.54 & 1.34 & 0.24 & 1.40 & 0.17    \\
        TSMixer      & 1.08 & 0.53 & 1.34 & 0.24 & 1.34 & 0.24    \\
        TimeMixer    & 1.39 & 0.23 & 1.41 & 0.16 & 1.45 & 0.11   \\
        PatchTST     & 1.25 & 0.37 & 1.34 & 0.24 & 1.38 & 0.19    \\
        DEUT         & 1.11 & 0.51 & 1.32 & 0.27 & 1.38 & 0.20   \\
        \textbf{CHAM-net}   & \textbf{0.92} & \textbf{0.66} & \textbf{1.16} & \textbf{0.43} & \textbf{1.25} & \textbf{0.34}    \\
        \bottomrule
    \end{tabular}
    \label{pretrained}
\end{table*}

\textbf{Transfer Learning Analysis}
We report the performance results of adopting transfer learning methods (e.g., pretraining and fine-tuning).
Table \ref{pretrained} reports the performance on observational datasets when models are first pretrained on the simulation datasets and then fine-tuned on observational data. 
In the table, the columns correspond to the simulation datasets used for pretraining, which are subsequently fine-tuned on the corresponding observational emission or consumption datasets.
For example, \teme denotes pretraining on \teme followed by fine-tuning on \fluxnete.

First, \cham consistently outperforms all baseline methods under the pretraining and fine-tuning setting, demonstrating its strong adaptability across data domains. 
Second, pretraining proves particularly beneficial for methane consumption tasks. 
As shown in Table \ref{pretrained}, pretraining on either \temc or MeMo improves both metrics. Notably, pretraining on \temc increases the R$^2$ value from 0.31 to 0.43 in \cham, indicating enhanced pattern learning from simulation datasets.
Finally, performance varies across different pretraining sources, suggesting that the quality of simulation datasets directly impacts fine-tuning effectiveness. Because \temc incorporates stronger physical constraints and has more realistic consumption dynamics and scales, it provides a more informative prior and leads to higher fine-tuned performance on \fluxnetc dataset.

\subsubsection{Model Analysis}

\begin{figure}[!h]
    \centering
    \includegraphics[width=0.9\linewidth]{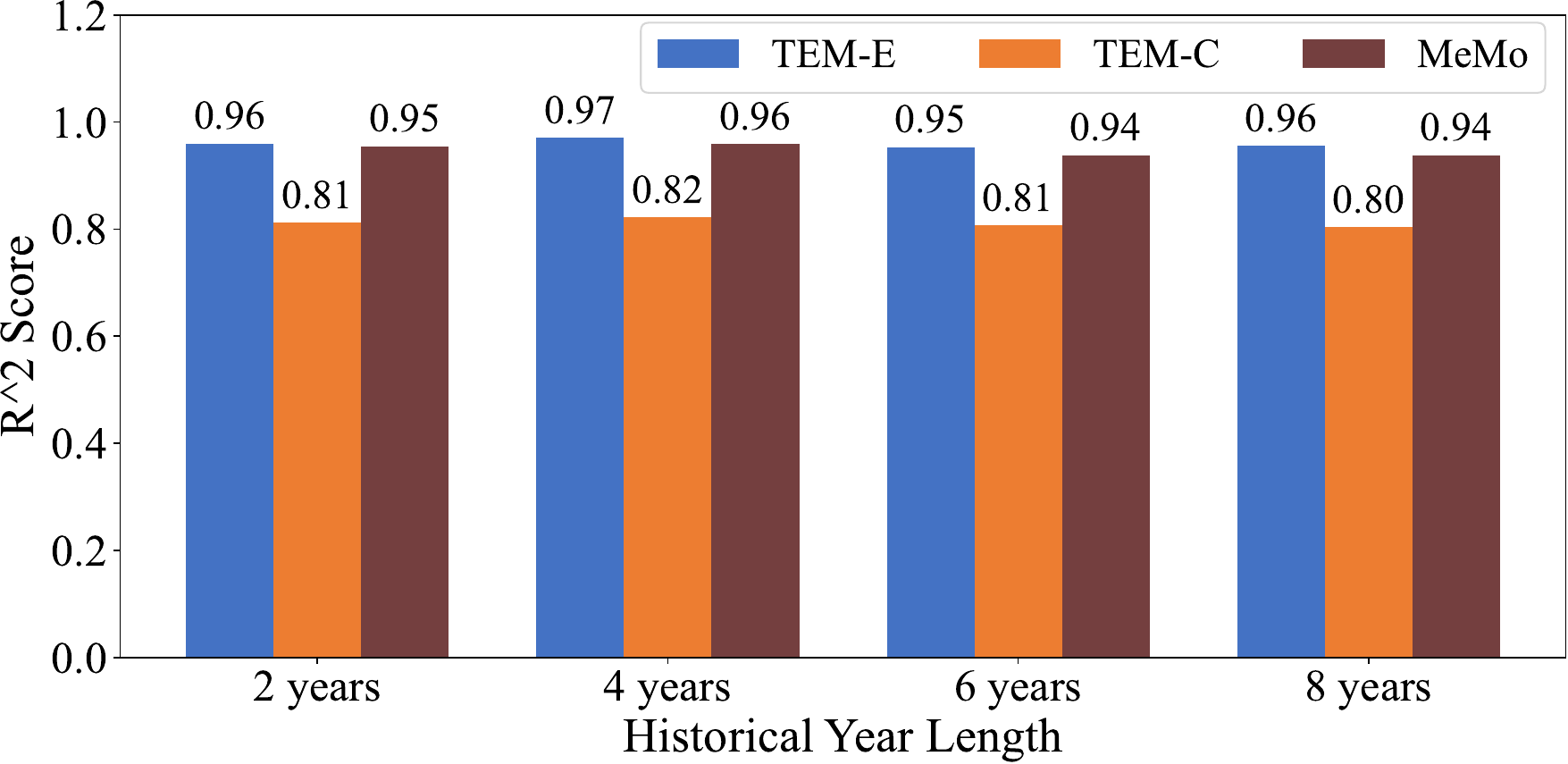}
    \vspace{-0.2cm}
    \caption{Sensitivity study of \cham with respect to historical year length.}
    \vspace{-0.4cm}
    \label{year_length}
\end{figure}

\textit{Historical Year Length}. We first investigate the optimal length of historical data used in the inner-loop of \cham. As shown in Figure \ref{year_length}, we report performance using the R$^2$ value to compare different historical year window lengths. 
We evaluate historical year lengths of 2, 4, 6, and 8 years to determine the optimal setting for the simulation datasets \teme, \temc, and MeMo.
Due to the sparsity and varying temporal spans of the observational datasets (e.g., some sites contain only three years of data), we use a single historical year length for \fluxnete and \fluxnetc datasets in our experiments.
This setting is sufficient to yield explicit performance improvements, as demonstrated in Table \ref{main_results}.
For the simulation datasets, Figure \ref{year_length} shows that performance varies marginally across different lengths of historical window.
Nevertheless, a four-year historical window consistently achieves the best performance. 

\textit{Learned Representations Analysis}. 
To interpret the learned representations and identify the most influential inputs, 
we analyze the correlation between the learned representations and the input variables. 
Using \fluxnete as an example, we extract the site-specific representations for all sites after training and apply principal component analysis (PCA) to identify the top three dominant components. We then compute the Pearson product–moment correlation coefficients between these components and the site-wise input features.

\begin{figure}[!h]
    \centering
    \includegraphics[width=1\linewidth]{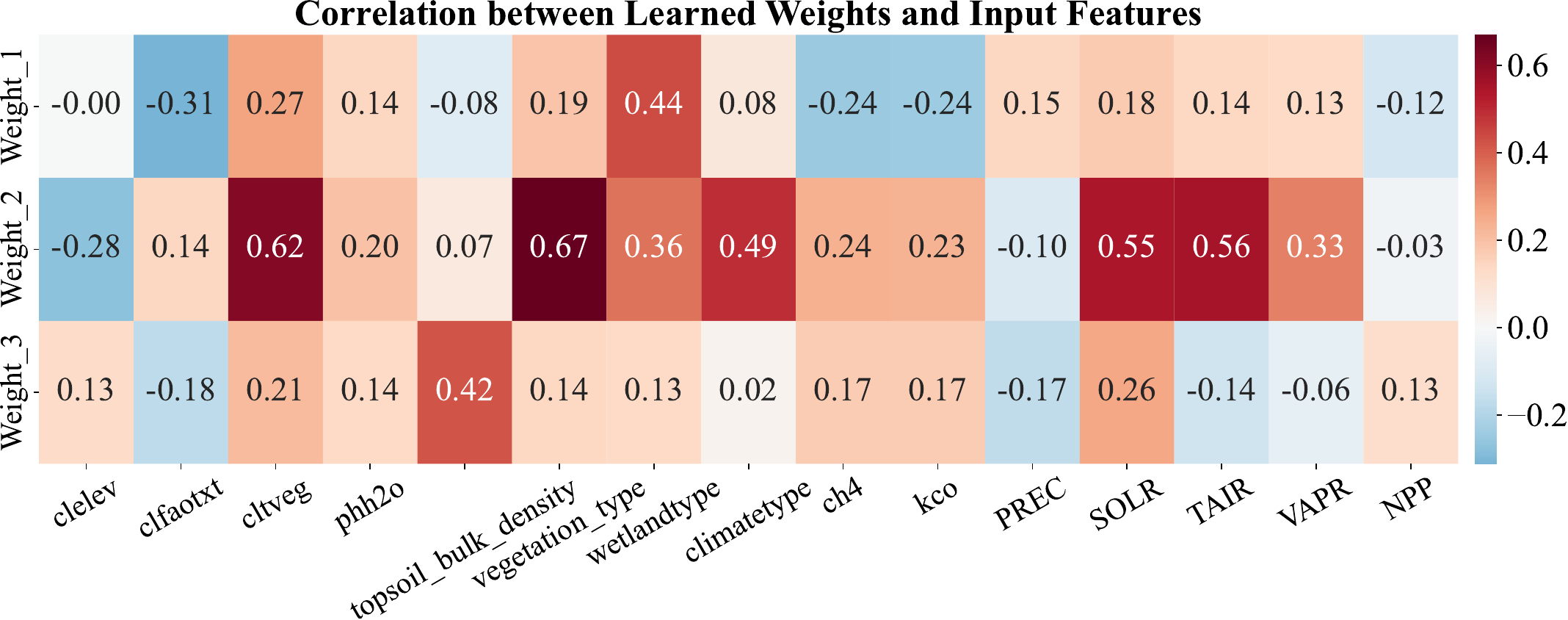}
    \vspace{-0.2cm}
    \caption{Correlation between learned weights and input features.}
    \vspace{-0.1cm}
    \label{weights_analysis}
\end{figure}

As shown in the heatmap in Figure \ref{weights_analysis}, the learned representations capture meaningful physical relationships when encoding historical information. 
In the figure, the x-axis represents the 15 input features, while the y-axis corresponds to the three most important components identified by PCA, namely Weight\_1 to Weight\_3. In the heatmap, 
positions in read are indicative of strong associations.
The results show that topsoil bulk density, vegetation types (\texttt{cltveg} and \texttt{vegetation\_type}), climate type, solar radiation (\texttt{SOLR}), and air temperature (\texttt{TAIR}) are among the most influential features. 
These findings are consistent with the established understanding of methane processes in natural ecosystems and suggest that the learned representations can provide useful insights for improved methane forecasting and mitigation strategies.
\vspace{-0.1cm}

\subsubsection{Ablation Studies}
We also conduct the ablation experiments to examine the contribution of each component in our framework. Using the emission datasets as an example, we evaluate three variants of \cham: (i) a mean-pooling-based \cham, (ii) \cham with cross-attention, and (iii) \cham with both cross-attention and contrastive learning. 
For reference, we also include the LSTM as baseline.

\begin{figure}[!h]
    \centering
    \includegraphics[width=0.85\linewidth]{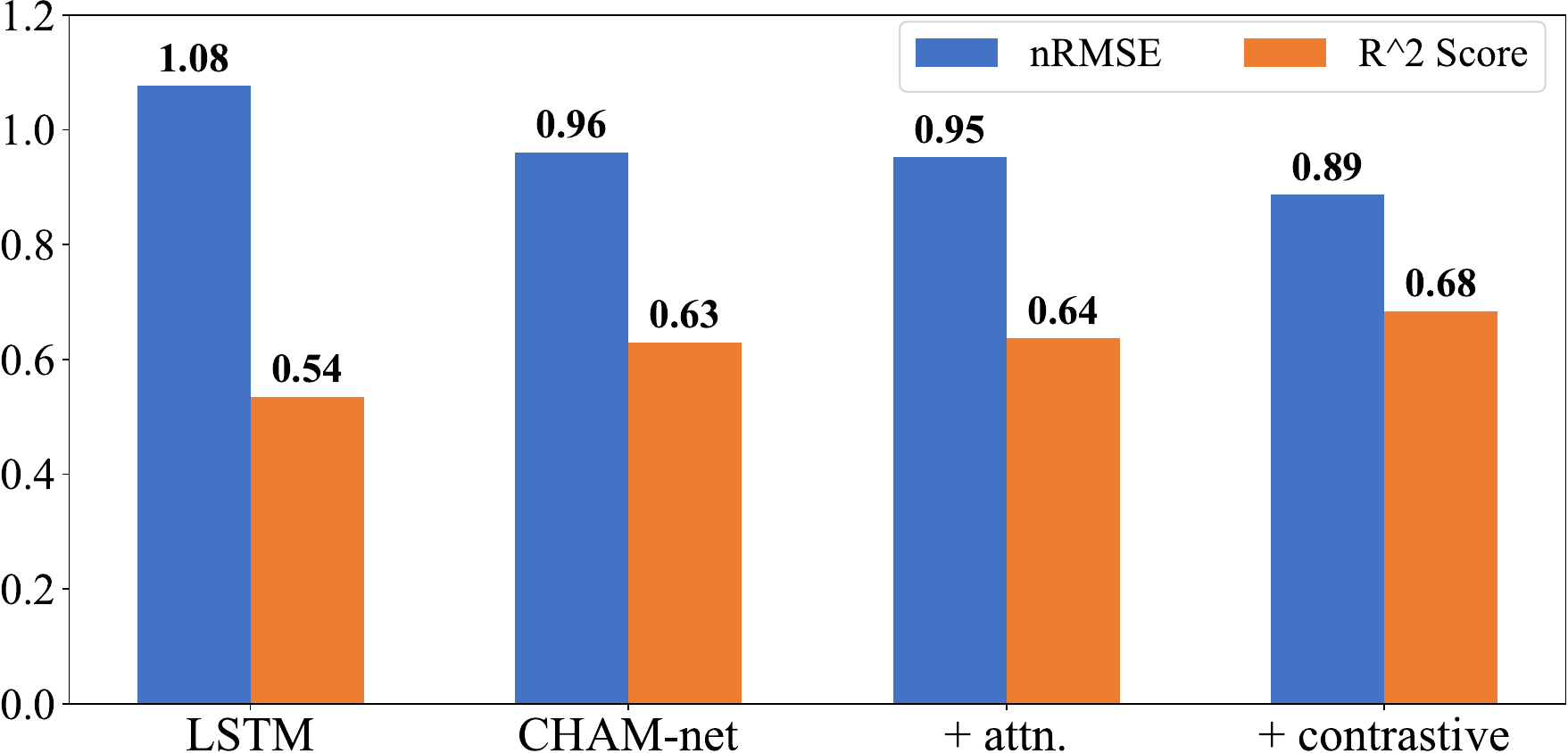}
    \vspace{-0.1cm}
    \caption{Ablation study of different model variants (in R$^2$).} 
    \vspace{-0.5cm}
    \label{ablation}
\end{figure}

As shown in the Figure \ref{ablation}, 
the incorporation of historical information accounts for the largest performance improvement. 
Contrastive learning provides the second-largest gain, as it enhances the model’s ability to distinguish site-specific characteristics and extract informative features from historical data, leading to more expressive embeddings.
Cross-attention yields a relatively smaller improvement in this setting. This is because input features evolve slowly over time, resulting in high similarity across historical years, thus the cross-attention offers limited benefit over mean pooling. 
Nevertheless, cross-attention remains a configurable component and may offer greater benefits when applied to other domains where inputs diversity across years is more pronounced.

\section{Related Work}
\paragraph{Methane Prediction.}
Previous methane flux prediction studies~\cite{ml2,ml1,ml3,ml5,ml4} primarily rely on Random Forest, decision tree, extreme gradient boosting (XGB), and Artificial Neural Network (ANN) approaches trained on relatively small observational datasets, which often focused on specific regions and coarse spatial resolutions. The work of~\cite{XMethaneWet} introduced the first global wetland methane emission dataset that integrates both simulation and observational data, enabling large-scale machine learning studies. 
To the best of our knowledge, this work is the first to jointly model methane emission and consumption in natural ecosystems using machine learning, achieving state-of-the-art performance and providing new insights for global methane budget estimation and analysis.

\paragraph{Knowledge-guided machine learning.} 
Knowledge-guided machine learning (KGML)~\cite{kgml1,kgml2,kgml5} has been successfully applied in various environmental studies, including carbon dioxide modeling~\cite{kgml3} and lake temperature profiling~\cite{kgml4,adptive_water}, by embedding physical knowledge into the loss functions of ML models to enhance performance. 
While directly incorporating biogeochemical equations into methane prediction models is beyond the scope of this work, it represents a promising direction for future research. In the supplementary material, we leverage pretraining and fine-tuning to transfer knowledge from simulation datasets to observational datasets, which can also be viewed as a form of knowledge guidance.

\section{Conclusion}
In this paper, we propose a contrastive hierarchical adaptive meta-learning framework that explicitly leverages site-specific historical information to capture both spatial heterogeneity and temporal dynamics in methane prediction.
By learning site-aware representations, our model improves prediction accuracy for both methane emission and consumption across simulation and observational datasets. These improvements support more reliable estimation of the global methane budget and provide insights that may inform effective strategies for natural methane mitigation.
Experimental results demonstrate that our approach consistently outperforms all other baselines, achieving the lowest nRMSE and the highest R$^2$ on both methane emission and consumption datasets.


\section*{Acknowledgements}
Rongchao Dong and Xiaowei Jia were partially supported by the National Science Foundation (NSF) grants 2203581, 2239175, 2316305, 2147195, 2425845, and 2530609; the USGS award G22AC00266; and the NASA grants 80NSSC24K1061 and 80NSSC25K0013.
Licheng Liu and Youmi Oh are supported by the Department of Energy (DOE) grant DE-SC0024360, NSF ESIIL 2153040.
Yiqun Xie is supported in part by the NSF under Grant No. 2126474, 2147195, 2425844, and 2530610; NASA under grant 80NSSC25K0013 and 80NSSC25K7221; Google’s AI for Social Good Impact Scholars program.
We also sincerely thank all reviewers for their thoughtful comments and feedback, and all our collaborators for their insightful contributions.

\bibliographystyle{named}
\bibliography{ijcai26}

\end{document}